# A Survey of Surface Defect Detection of Industrial Products Based on A Small Number of Labeled Data

Qifan Jin (1), Li Chen (1 and 2)

((1) College of Computer and Artificial Intelligence, Zhengzhou University, (2) Institute of Physical Education (Main Campus), Zhengzhou University)

**Abstract:** The surface defect detection method based on visual perception has been widely used in industrial quality inspection. Because defect data are not easy to obtain and the annotation of a large number of defect data will waste a lot of manpower and material resources. Therefore, this paper reviews the methods of surface defect detection of industrial products based on a small number of labeled data, and this method is divided into traditional image processing-based industrial product surface defect detection methods and deep learning-based industrial product surface defect detection methods suitable for a small number of labeled data. The traditional image processing-based industrial product surface defect detection methods are divided into statistical methods, spectral methods and model methods. Deep learning-based industrial product surface defect detection methods suitable for a small number of labeled data are divided into based on data augmentation, based on transfer learning, model-based fine-tuning, semi-supervised, weak supervised and unsupervised.

# 1 Introduction

In industrial production, product quality inspection is particularly important. At present, the most commonly used method for surface defect detection of industrial products is mainly manual inspection. However, manual defect inspection is easily affected by the subjective factors and work experience of the inspector, and the accuracy rate is not high. When the defect is small, the human eye cannot detect it, and it's not suitable for manual inspection when some inspection environments are harmful to the human body. In recent years, surface defect detection of industrial products based on visual perception has gradually replaced manual defect detection and become an indispensable part of industrial product surface defect detection. Surface defect detection of industrial products based on visual perception technology is a non-contact automatic detection technology. It has high accuracy and high accuracy, and can run for a long time in a complex production environment, it is widely used in tile defects [2][21], fabric defects [3][18][23][41][53][68], steel plate defects[5][11][15][32][55][67], PCB defects [54][60][65] and other directions. The widespread application of surface defect detection of industrial products based on visual perception not only reduces labor costs, but also significantly improves production efficiency.

Surface defect detection of industrial products based on visual perception can be mainly divided into traditional image processing-based industrial product surface defect detection and deep learning-based industrial product surface defect detection. Traditional image processing-based industrial product surface defect detection requires manually designed features. This method of manual feature design avoids using a large number of data to learn features, but it has stricter requirements on the imaging environment and poor adaptability. Deep learning-based industrial product surface defect detection needs to automatically extract features from a large number of data. This method of automatic feature extraction is highly adaptable to the environment, but requires a large number of training data. In the real industrial environment, due to the low probability of occurrence of defective samples, it will be difficult to collect defective samples, and defect labeling requires a lot of manpower and material resources. Therefore, surface defect detection of industrial products based on a small number of labeled data is particularly important.

This paper introduces the method of surface defect detection of industrial products based on a small number of labeled data, and the method is roughly divided into traditional image processing-based industrial product surface defect detection method and deep learning-based industrial product surface defect detection methods suitable for a small number of labeled data. The structure of this paper is as follows. Section 2 introduces traditional image processing-based industrial product surface defect detection methods, which are divided into statistical methods, spectral methods and model methods. Section 3 introduces deep learning-based industrial product surface defect detection methods suitable for a small number of labeled data, which are divided into based on data augmentation, based on transfer learning, model-based fine-tuning, semi-supervised, weak supervised and unsupervised. Section 4 concludes.

## 2. Traditional Image Processing-Based Industrial Product Surface Defect Detection

Traditional image processing-based surface defect detection of industrial products needs to design a method for extracting image features according to the texture features of the image, which does not require a large number of labeled data to train the model. This section divides such methods into statistical methods, spectral method and model methods. This section will introduce them separately.

### 2.1 Statistics Method

Statistical method is a method to analyze texture and identify defects according to the statistical distribution characteristics of images. This method can be further divided into grayscale statistical method, pattern statistical method and gradient statistical method. The description of texture features by statistical methods is a point-by-point calculation process, which is basically unaffected by illumination and noise, and has strong stability. However, because it cannot obtain global information, it is impossible to study the dependencies between pixels.

#### 2.1.1 Grayscale Statistics

The grayscale statistics method is a method to obtain the texture features of the image by counting the grayscale values of the image pixels. The commonly used grayscale statistics feature methods are mean, variance, histogram and so on. BRADSHAW [1] used the gray mean and variance of the image to be detected to calculate two thresholds above and below the mean, and used these two thresholds to detect the surface defects of light-colored fabrics and dark-colored fabrics. STOJANOVIC [2] proposed a fabric detection system based on improved binary, texture and neural network algorithms, which achieved good results in detecting various types of fabric defects. TSAI [3] proposed a fast regularity metric for defect detection in non-textured and uniformly textured surfaces. The method first selects an appropriately sized neighborhood window and builds a weighted variance matrix. Finally, identify defects based on similarity of eigenvalues of weighted variance matrix. NAND [4] proposed a new entropy-based defect detection algorithm. This method proposes a concept of dynamic update, which helps to find an ideal background, then calculates the difference between the local entropy of the image to be detected and the background image, and then identifies defects according to the histogram feature of the entropy difference. YANG [5] proposed a detection method based on histogram differences. From the perspective of image segmentation, taking the defect area as the foreground and the product surface of the non-defect area as the background, the difference of the gray distribution histogram of the two areas is used as the main segmentation basis.

**2.1.2 Pattern Statistics**

Gray-Level Co-occurrence Matrices (GLCM) and Local Binary Patterns (LBP) are commonly used statistical features of patterns. HARALICK [6] first proposed the gray-level co-occurrence matrix, which defines a total of 14 texture features. In order to reduce the time used for feature extraction, LI [7] fused the sub-images obtained after Gabor transformation and extracted texture features, and then used the improved Relief algorithm to reduce the feature dimension. In order to improve the detection accuracy, bias processing is introduced and a least squares support vector machine is used for classification. SIEW [8] found that some features extracted from the spatial grayscale dependency matrix, the adjacent grayscale dependency matrix, the grayscale difference method and the grayscale run-length method can objectively identify the carpet texture and its changes during the wear process. OJALA [9] first proposed LBP and used LBP for texture image classification. IIVRINEN [10] compared the two feature extraction methods of the gray-scale co-occurrence matrix and the local binary pattern through experiments. The experiments show that there is no obvious difference between the two feature sets in the experimental results, but the gray level co-occurrence matrix is less computationally efficient than the local binary pattern. SONG [11] proposed a robust feature descriptor based on the fully local binary patterns, called the adjacent evaluation completed local binary patterns (AECLBPs), which constructs an adjacent evaluation window around the neighbors. Through this window, the threshold value of the fully local binary patterns is modified, and the feature extraction method is applied to the defect detection of hot-rolled steel strip. WANG [12] proposed an Entity Sparsity Pursuit (ESP) method to identify surface defects. In terms of feature extraction, an intuitive LBP-inspired feature extractor is designed for industrial grayscale images. The extracted features can simultaneously analyze the grayscale distribution of domain pixels in the horizontal, vertical and two diagonal directions.

### 2.1.3 Gradient Statistics

Gradients characterize the changes in the grayscale of pixels in an image. Locations with high gradients and large grayscale changes are called edges. In texture images, the number of edges within a unit area is an important feature. ROBERT [13] designed a real-time defect detection system based on an edge detection algorithm. SONG [14] proposed a new algorithm to solve the problem using color and texture information and developed a new color clustering scheme based on human color perception. It is suitable for random texture images and produces good results especially in defect detection of granite images. WANG [15] proposed an improved random forest algorithm with optimal multi-feature-set fusion (OMFF-RF algorithm), and used it for defect recognition, OMFF-RP combines HOG feature set and GLCM feature set through multi-feature-set fusion factor to describe local and global texture features and then uses random forest algorithm for defect classification. TAN [16] proposed a fusion algorithm based on tile contour features and grayscale similarity matching, which realized the contour extraction of insulator tiles and the precise separation of insulator tiles, and they constructed an insulator defect detection model based on tile spacing and grayscale similarity matching. LI [17] proposed an efficient fabric defect detection algorithm based on histogram of oriented gradients (HOG) and low-rank decomposition. First, the fabric image is divided into image blocks of the same size, the HOG features of each image block are extracted, and the image block features are formed into a feature matrix, and an effective low-rank decomposition model is constructed for the feature matrix, and the solution is optimized by the Alternating Direction Methods (ADM) to generate low-rank and sparse arrays. Finally, the improved optimal threshold segmentation algorithm is used to segment the saliency map generated by the sparse matrix, so as to locate the defect area. The experimental results show that the low-rank decomposition can effectively realize the rapid separation of fabric defects. Compared with the existing methods, this method can significantly improve the defect detection performance of complex fabric texture images.

### 2.2 Spectral Method

For some defects that are difficult to be effectively identified only by pixel grayscale, the researchers believe that the appearance of defects destroys the structural consistency of uniform textures. We can transform the image in time domain. After the time domain transformation, accurate detection of defects can be achieved according to the different responses of normal texture and defect texture. This kind of method can be subdivided into Fourier Transform (FT), Gabor filter and wavelet transform. Fourier transform realizes the separation and conversion from time domain to frequency domain, and it is suitable for deterministic and stationary signals. It belongs to global analysis and is especially suitable for analyzing relatively stable signals for a long time. Gabor transform realizes the localization of time-frequency analysis and is a time-frequency analysis with a single resolution, which is especially sensitive to image edges and insensitive to illumination. Wavelet transform has excellent localized video analysis characteristics. It can change the spectral structure and window shape, and is very effective for analyzing mutation signals and singular signals.

## 2.2.1 Fourier Transform

Fourier transform transforms the grayscale change distribution function of the image into the frequency distribution function of the image, and the image can be processed in the obtained frequency domain as needed. Low-pass filtering will make the image blurred, it can filter out the noise of the image. High-pass filtering obtains the edge and texture information of the image. By enhancing the high-frequency signal of the image, the contrast of the image can be enhanced. Ciamberlini [18] developed and tested an optical configuration for fault detection based on the optical Fourier transform and applied it to fabric defect detection. TSAI [19] proposed a global image restoration scheme based on Fourier transform. First, the original image was transformed into a frequency domain image by Fourier transform, then use a one-dimensional Hough transform to detect the high frequency components in the frequency domain image, set them to zero, and finally inverse transform to a spatial domain image. In the recovered image, the uniform line areas in the original image will have approximately uniform gray levels, while the defect areas will be clearly preserved. Experiments demonstrate the effectiveness of the proposed method on various textile fabrics, machined surfaces, and natural woods. AI [20] proposed a multi-scale feature extraction method based on Curvelet transform and Kernel Locality Preserving Projections (KLPP). Curvelet transform is a non-adaptive image representation that does not need to know the geometric characteristics of the image. It is a multi-scale geometric analysis method with two important characteristics: anisotropic scaling law and directionality. Therefore, the Curvelet transform is suitable for describing and analyzing edge features of images with curvilinear or linear shapes. Since too many features using Curvelet transform will lead to a decrease in efficiency or a dimensional disaster, redundant features should be removed and important features should be retained. KLPP is a nonlinear dimensionality reduction method, which can well preserve the local neighborhood information of high-dimensional datasets and effectively eliminate the influence of noise and outliers. The method is used for feature extraction of multi-scale and multi-directional defect images, and then the support vector machine (SVM) is used for classification. WANG [21] constructed an image-based magnet surface defect detection system and proposed a Fourier image reconstruction method based on the magnet surface image processing method. A spectral image of the magnet image was obtained using a Fourier transform, and the defect is shown as a bright line in the image. Use the Hough transform to detect the angle of the bright line, then remove this line to remove the defects in the original grayscale image, and finally apply the inverse Fourier transform to get the background grayscale image. Defect areas are obtained by evaluating the grayscale difference between the original image and the background grayscale image. The experimental results show that the method can automatically and efficiently detect cutting defects on the surface of magnets. Zori [22] proposed a novel Fourier spectral loop feature extraction method for the defect detection problem of biscuit bricks, which creates feature vectors based on a two-dimensional Fourier power spectrum.

## 2.2.2 Gabor Filter

The Gabor filter can extract the texture information of the image in various scales and directions, and at the same time reduces the influence of illumination changes and noise in the image to a certain extent. BODNAROVA [23] designed an optimal Gabor filter to segment fabric surface defects. BODNAROVA [24] successfully designed a two-dimensional optimal Gabor filter by automatically and adaptively selecting the optimal parameters. Although only a small number of optimal filters are constructed, the computation time is greatly shortened, thereby increasing the detection speed. Doo-Chul [25] proposed a method for pinhole detection in billet images based on Gabor filtering and morphological features. First, Gabor filter was used to extract defective candidates, and then morphological features were used to identify pinholes in defective candidates. ZUNIGA [26] proposed a new method to enhance the Gabor wavelet process to extract fractal features in magnitude space. Experimental results on multiple texture image databases show that this method outperforms other earlier proposed methods and creates a more reliable technique for texture feature extraction. ZHANG [27] proposed an efficient method to detect fabric defects. This method utilizes Optimal Gabor Filter and Binary Random Drift Particle Swarm (BRDPSO) algorithm, which can simultaneously realize feature selection and parameter optimization. The parameters of the two-dimensional Gabor filter are adjusted by Quantum-behaved Particle Swarm Optimization (QPSO) to obtain the optimal Gabor filter. BRDPSO is used to select features on the original feature set while optimizing the parameters of the Isolation Forest (IF) classifier. A large number of experimental results show that the method has effective detection performance for defect detection of textile fabrics. ZHAO [28] proposed a method for defect detection of thermoelectric cooler components, which uses a Gabor filter to extract texture information in thermoelectric cooler component images as classification features. Meanwhile, Principal Component Analysis (PCA) was used to select categorical features. For the category of defects, SVM is used to classify unknown defects. By extracting features from thermoelectric cooler component images and using them for SVM training. Finally, taking the unknown defect image as input. The defect category is obtained. The superiority of this method is proved by experiments.

**2.2.3 Wavelet Transform**

Wavelet transform has very important applications in image processing, including image compression, image denoising, image fusion, image decomposition, image enhancement and so on. Wavelet transform has the advantages of low entropy, multi-resolution characteristics, decorrelation and flexibility of base selection. JASPER [29] applied Sobel edge operator, fast Fourier transform and discrete wavelet transform to fabric defect detection with missing wefts. Experiments show that the wavelet transform can characterize defects due to lack of pick and end faster and more accurately than other methods. JASPER [30] proposed a new method to capture texture information using an adaptive wavelet basis. The adaptive wavelet basis is very sensitive to the abrupt changes in texture structure caused by defects. SARI-SARRAF [31] proposed a defect segmentation algorithm based on wavelet transform and image fusion, which can enhance defect features, eliminate the influence of background regions, and it achieves a false alarm rate of 2.5% for fabric defects. Kim [32] introduced a vision-based online detection method for woven fabrics, using wavelet transform to extract features with specific defect features, and experimentally demonstrated the effectiveness of the proposed detection system on different types and styles of fabrics sex. GHORAI [33] proved that the wavelet feature set is more suitable for the application of steel surface defect detection compared with the techniques based on texture segmentation and thresholding. In addition, five different wavelet features are compared in the detection of steel surface defects, and it is concluded that the effect of Haar wavelet features is better. Sulochan [34] proposed a weave detection method based on multi-scale wavelet features and fuzzy clustering method, which uses fuzzy C-means clustering algorithm to detect two intersections in weave patterns from multi-scale wavelet features regional status. The texture orientation feature of the intersection area determines the exact state of the intersection area. A binary pattern matrix is obtained from the detected intersection region states. Finally, an error correction method is used to correct the errors in some intersecting regions, and a corrected pattern is obtained. The method is validated using computer simulated woven fabric samples and real woven fabric images.

**2.3 Model Method**

Model-based texture analysis methods attempt to capture the process of generating textures. They try to model textures by determining the parameters of predefined models. The model method not only considers the local randomness of the texture, but also considers the regularity of the whole texture, which is more flexible. but it is difficult to estimate the parameters of the model, and the number of calculation is large. COHEN [35] used Gaussian Markov random fields to model texture images of defect-free fabrics. MAO [36] proposed a Multiresolution Simultaneous Autoregressive (MR-SAR) model for texture classification and segmentation. First, a multivariate Rotation-invariant Simultaneous Autoregressive (RISAR) model based on the Circular Autoregressive (CAR) model is introduced. Experiments show that the multivariate RISAR model is superior to the CAR model in texture classification. BAYKUT [37] studied various texture analysis algorithms for textile quality inspection, and presented a parallel implementation design of the algorithm for Markov random field modeling based on TMS320C40 system. XIE [38] proposed a texture paradigm model for defect detection and localization in random color textures. XU [39] divided the surface defect images of hot-rolled steel strips into two categories: "background" and "defects", and used Hidden Markov Tree (HMT) models to model and achieve multi-scale defect segmentation respectively. In view of the high segmentation error rate in the fine-scale segmentation results obtained by the HMT model, A context-adaptive hidden Markov tree (CAHMT) method is introduced to fuse the segmentation results of different scales, which greatly reduces the segmentation error rate of fine-scale segmentation.

# 3. Deep Learning-Based Industrial Product Surface Defect Detection

Traditional deep learning-based surface defect detection requires a large number of data to learn features. In real industrial environments, defect data is not easy to obtain and labeling data requires a lot of manpower and material resources. Therefore, deep learning-based industrial product surface defect detection methods suitable for a small number of labeled data is introduced in this section.

## 3.1 Based on Data Augmentation

In the case of limited labeled data, data augmentation can be used to improve the diversity of samples. Data augmentation allows limited data to generate value equivalent to the value of more data without substantially increasing data. Data augmentation uses auxiliary data or auxiliary information to perform data augmentation or feature enhancement on a small number of original annotated samples.

### 3.1.1 Methods Based on Data Synthesis

The early data synthesis method is to generate new data by translating, rotating, deforming, scaling, color space transformation, cropping and other methods of existing data. At present, there are many more effective data synthesis methods, such as generating Generative Adversarial Networks (GAN), artificially generated defect data, etc. Liu [40] proposed a GAN-based simulation method for general defect samples. Under the GAN framework, a simulation network with an encoder-decoder structure was proposed. Furthermore, the simulated network and the discriminative network are adversarially trained under the proposed region training strategy, which prioritizes the translation of defective regions. Finally, defect-free regions are refined by wavelet fusion. The method requires a small number of defective training samples and can generate simulated defects of specified shape and type. WANG [41] proposed a true and false data fusion algorithm based on a deep convolutional confrontation generation network and random image stitching, aiming at the problems of network overfitting and substandard performance of the model caused by the lack of solar cell defect data. The number of training data is increased by 800 times, and the network model is lightweight and optimized to reduce model training parameters. Research has proved that the true and false data fusion algorithm can effectively alleviate the problem of network overfitting caused by insufficient training data. While ensuring the accuracy of the lightweight optimization model, it can compress the model size and speed up the test speed. Wei [42] proposed a new method suitable for small sample fabric defect detection based on compressed sensing and Convolutional Neural Networks (CNN). The method utilizes the compressed sampling theorem to compress and expand the data with small sample size, and then uses CNN to classify the data directly obtained from the compressed sampling. The experimental results show that the method can effectively improve the classification accuracy of fabric defect samples even when the number of defect samples is small. Haselmann [43] proposed a method to synthesize defects, which can be used for dataset expansion for surface defect detection. The entire synthesis process mainly includes four steps: 1. skeleton generation, 2. texture generation, 3. modification of defect-free images, 4. analysis of defect visualization degree. Li [44] proposed a synthetic defect algorithm. First, four typical collection screen defects are simulated by the algorithm, and an artificial defect database was created, then the artificial data set was applied to the deep learning recognition algorithm, and the initial model was trained. Then the initial model is fine-tuned and retrained with real defective samples. Finally, use the retrained model to test real samples. Tao [45] proposed a novel cascaded structure of deep convolutional neural networks for localization and detection of defects in insulators. The cascade network transforms the defect detection problem into a two-level object detection problem using a region proposal network-based CNN. To address the scarcity of defective images in practical detection environments, a data augmentation method is proposed, which includes four operations: 1. affine transformation, 2. insulator segmentation and background fusion, 3. Gaussian blur, 4. luminance transformation. The experimental results show that the method meets the robustness and accuracy requirements of insulator defect detection. Sun [46] proposed a novel surface defect detection algorithm based on Adaptive Multiscale Image Collection (AMIC) of convolutional neural network. First, the inspection network is pre-trained on the ImageNet dataset. Second, AMIC is established, which consists of adaptive multi-scale image extraction and contour local extraction of training images. With AMIC, the training dataset can be greatly increased, and image labeling can be done automatically. Then, transfer learning is performed using the AMIC established from the training dataset. Finally, an embedded metal surface defect automatic detector is designed, and a detection algorithm is proposed.

### 3.1.2 Method based on feature enhancement

The key to surface defect detection with a small number of labeled data is to obtain a feature extractor with better generalization. Feature enhancement refers to enhancing the features that are easy to classify in the original sample space and weakening the invalid features, so as to improve the diversity of sample features. Lv [47] proposed a novel few-shot learning method combined with an attention mechanism, which is constructed by a CNN that extracts image features and a Relation Network (RN) that computes the similarity score between a pair of images, image categories can be predicted from similarity scores. In order to extract more effective and discriminative features, Squeeze-and-Excitation Networks (SENet) are introduced in this method as an attention module to enhance effective features and weaken ineffective features.

### 3.2 Based on Transfer Learning

Transfer learning can retrieve shared knowledge from the source task and apply it to the learning of the target task, thereby improving the generalization performance of the model and reducing the time and resource consumption caused by the large number of labeled data required by the target task. Ren [48] proposed a general automatic surface detection method that requires a small number of training data. The method constructs a classifier based on the features of image patches, where the features are transferred from a pretrained deep learning network, and then obtains pixel-wise predictions by convolving the trained classifier with the input image. Ferguson [49] proposed a casting defect recognition system in X-ray images based on the mask region CNN structure. The defect detection system simultaneously performs defect detection and segmentation on the input image, and it is suitable for a variety of defect detection tasks. The model uses transfer learning to reduce training data requirements and improve the prediction accuracy of the trained model. More specifically, the model is first trained using two large datasets of publicly available images, and then the trained network is transferred to a relatively small dataset of metal casting X-rays. Gong [50] proposed a transfer learning object detection based on Domain Adaptive (The purpose of pre-adaptation is to make the probability distribution of the source and target domains as the same as possible) Faster R-CNN (DA Faster) for the case where the defect size in the X-ray images of the spacecraft composite structure is very small and the number of training samples is small Model. On the basis of DA Faster, a Feature Pyramid Network (FPN) is added to the feature extraction sub-module for multi-scale feature adaptation, and the small anchor strategy and ROI Align are used to help locate small-size defects. HUANG [51] introduced the metric learning method in small-sample learning into the field of defect detection, and they proposed a small-sample metric transfer learning method to solve the problem of requiring a large number of learning samples in deep learning methods. Experiments show that small-sample metric learning enables the network to obtain better performance with fewer samples by learning on uncorrelated large-scale datasets. LIU [52] proposed a knowledge reuse strategy to train a CNN model to improve the accuracy and robustness of defect detection. By introducing model-based transfer learning and data augmentation, knowledge from other vision tasks is transferred to the industrial defect detection task to achieve high accuracy with limited training samples.

### 3.3 Model-based fine-tuning

Model fine-tuning first pre-trains on a large-scale source data set, and then uses a small number of labeled target data sets to fine-tune the parameters of the fully connected layer or top layers of the neural network model to obtain a fine-tuned model. When the distribution of the target dataset with a few annotations is similar to the source dataset, the effect of model fine-tuning is better. For optical detection of textured images. Kim [53] transfer weights using a source network trained on arbitrary unrelated images in the ImageNet dataset. Test experiments performed by this method show that one fine-tuning is sufficient to achieve a classification accuracy of 99.95%. An in-depth analysis of the effects of fine-tuning shows that most of the unnecessary features in the source network are deactivated after fine-tuning, while the meaningful features of the target data are amplified to capture new changes in the target domain. JING [54] proposed a fabric surface defect classification method based on convolutional neural network. Due to the large number of parameters and large sample size during CNN training, and it is easy to fall into overfitting, the fine-tuned convolutional neural network model Alexnet is used to extract features from the fabric defect image, and the parameters of the original network are used for initialization instead of random initialization parameters. The training samples under the target fine-tune the network parameters, and finally use the softmax regression algorithm for prediction and classification. ZHANG [55] proposed an improved PCB defect detection method by learning deep discriminative features, which greatly reduces the high requirements of deep learning methods for large datasets. The existing PCB defect dataset is extended with some artificial defect data and affine transformations to increase the quantity and diversity of defect data. Then, a deep pretrained convolutional neural network is used to learn high-level discriminative features of defects. Fine-tune the base model on the extended dataset by training the top layers. Finally, a sliding window method is used to further locate the defects. Experiments show that this method is more feasible and effective in the field of PCB defect detection.

### 3.4 Semi-supervised

Semi-supervised is to use a large number of unlabeled data and a small number of labeled data to train a surface defect detection model. Di [56] proposed a semi-supervised method based on Convolutional Autoencoder (CAE) and Semi-supervised Generative Adversarial Networks

(SGAN) to classify steel surface defects. The algorithm first trains CAE through a large number of unlabeled data, so that CAE has preliminary feature extraction capabilities. After the CAE training is completed, a GAN network is introduced, and the encoder part of CAE is transformed into a GAN discriminator, then using the labeled samples train the model. He [57] proposed a defect classification method based on multi-trained semi-supervised learning of two different networks for surface defect detection of industrial steel products. The method first trains the cDCGAN on the original labeled samples to generate enough unlabeled samples. Next, the residual network also performs initial training on the original samples and makes predictions on the GAN samples. Then, high-confidence samples are labeled with the estimated class based on the class confidence predicted by each classifier and added to the training set. Finally, the residual network is retrained on the new training set until all GAN samples have been added to training. Experimental results show that the method can achieve competitive classification performance even with limited raw samples. Gao [58] proposed a CNN-based semi-supervised learning method for steel surface defect recognition, and CNN was improved by Pseudo-Label (PL), an efficient semi-supervised framework that can be used for unlabeled of samples to generate false labels. The experimental results on the benchmark dataset for steel surface defect recognition show that the method can achieve good recognition results with limited labeled data. XIE [59] constructed a semi-supervised Deep Convolutional Generative Adversarial Network (DCGAN) model. The model first uses HSV (Hue Saturation Value) color space conversion and Otsu algorithm (Otsu) to preprocess the original injection bottle image to obtain a training set. Then the learning tasks are combined so that the unsupervised discriminator of DCGAN shares the convolutional layer parameters with the supervised classifier for surface defect detection of injection-molded bottles, and the loss function is modified at the same time. Adding cross-entropy to the Wasserstein distance of the DCGAN model, and finally using the Adam optimizer for model training. ZHANG [60] proposed a Semi-supervised Generative Adversarial Network (SSGAN) with two sub-networks to obtain more accurate segmentation results at the pixel level. One is a segmentation network for defect segmentation from labeled and unlabeled images, which is built on a dual attention mechanism. Specifically, the attention mechanism is used to extract rich and global representations of pixels in both spatial and channel dimensions for better feature representation. The other is a Fully Convolutional Discriminator (FCD) network, which employs two loss functions (adversarial loss and cross-entropy loss) to generate confidential density maps of unlabeled images in a semi-supervised learning fashion. Compared to most existing methods that rely heavily on labeled or weakly labeled images, the developed SSGAN model can leverage unlabeled images to enhance segmentation performance and alleviate the data labeling task. The effectiveness of the proposed SSGAN model is demonstrated in a public dataset with four classes of steel defects.

### 3.5 Weak Supervision

Weak supervision uses data with unreliable labels to train surface defect detection models. ZHANG [61] proposed a weakly supervised learning method named Category-Aware Object Detection Network (CADN). The CADN is trained using only image label annotations and performs both image classification and defect localization. Weakly supervised learning is achieved by extracting category-aware spatial information in the classification pipeline. CADNs can be equipped with lighter or larger backbone networks as feature extractors for better real-time performance or higher accuracy, employing a knowledge distillation strategy to force the learned features of lighter CADNs to mimic those of larger CADNs. Therefore, the accuracy of lighter CADNs is improved while maintaining high real-time performance. Experiments show that the proposed method achieves satisfactory performance, which can fully meet the industrial requirements. He [62] proposed a new weakly supervised deep learning based method to accurately segment and localize defects on textured surfaces. In the proposed method, one encoder for feature extraction and two decoders for two related tasks are designed. To ensure that deep neural networks benefit from multi-task learning, the two decoders share a unique encoder. The main task aims to recover defects on textured surfaces, and the auxiliary task aims to obtain a Region of Interest (ROI), which is used to filter out noise in the main task. Subsequently, a residual map can be obtained by comparing the original and restored images. Finally, more accurate results can be obtained by fusing residual maps and ROIs. A series of experiments on the public defect detection dataset DAGM show that the method exhibits the best performance compared to other state-of-the-art methods. CHEN [63] proposed a general CNN-based defect detection framework with an attention architecture and a random forest classifier. By fusing CNN with random forest classifier and spatial attention module, a new deep CNN model is proposed to solve the problem of defect classification in different surface textures, the new CNN model significantly improves the proposed CNN model classification effect and robustness. Combining attention mechanism and CAM, a robust Spatial Attention Class Activation Map (SACAM) network structure is designed. SA-CAM suppresses complex backgrounds with different textures while highlighting defect regions, which helps to generate more accurate saliency maps. The proposed weakly supervised learning segmentation method is based on the saliency map of SA-CAM. It uses global image labels to achieve pixel-level defect segmentation, which simplifies the pixel-level labeling task of complex surface defects and exhibits good generality for different textured surfaces.

### 3.6 Unsupervised

Unsupervised is to use only normal data or unlabeled data to train surface defect detection models. Volkau [64] proposed a one-class training model aimed at extracting unique semantic features from normal samples in an unsupervised manner. This paper proposes a variant of transfer learning that includes a combination of unsupervised learning used on VGG16 and pre-training on ImageNet weight coefficients. MEI [65] combined the idea of image pyramid hierarchy and Convolutional Denoising Autoencoder (CDAE) network to detect texture image defects. Image patches are reconstructed using convolutional denoising autoencoder networks with different Gaussian pyramid levels, the reconstruction residuals of the training blocks are used as indicators for direct pixel-wise defect prediction, and the reconstructed residual maps generated by each channel are combined. Generate the final detection result. LIU [66] proposed a new Haar-Weibull-variance (HWV) model for detecting steel surface defects in an unsupervised manner. Defect areas can be located by an adaptive threshold, which is calculated after evaluating the distance of each sample from the origin in M-HWV space. Experimental results show that the method can detect any type of defects on uniformly textured surfaces, with an average detection rate of 96.2% on the dataset, outperforming previous methods. Hu [67] proposed an unsupervised method based on the Deep Convolutional Generative Adversarial Network (DCGAN) model to inspect the surface defects of woven fabrics. In addition to the discriminator and generator models in standard DCGAN, this method adds a third ConvNet model to map any query data from the image space back to the latent space. Combining the inverter and generator can reconstruct a given query image, and by subtracting the reconstructed image from the original image, a residual map can be created to highlight defective areas while attenuating non-defective areas. To reduce the effect of noise that may be present in the residual map, the method applies a discriminator to generate a likelihood map that measures the probability that any local image patch is defective. The residual and likelihood maps are further synthesized together to form an enhanced fusion map. Typically, fusion maps exhibit uniform gray levels on defect-free areas and significant deviations on defect areas. Therefore, it provides a good cluster representation of uniformly textured regions and can be used as a metric to distinguish defects from background. Potential defects can be identified and separated by simple thresholding on the fusion map. Dong [68] proposed a novel unsupervised local deep feature learning method based on image segmentation, and used this method to build a network that can extract useful features from images. The method uses pseudo-labels assigned by k-means clustering to train the segmentation U-Net and proposes an alternating algorithm in which the clustering and segmentation training steps are interleaved. In this case, CNN training and clustering can help each other. He [69] proposed an unsupervised defect detection algorithm based on cascaded GANs with edge repair feature fusion. In this algorithm, the edge repair network provides the defect repair network with complete structural features through a feature fusion method based on channel attention. For edge inpainting networks, a deformable autoencoder is developed that fully exploits the ability of deformable convolutions to perceive very little contextual information to improve its ability to inpaint defective edges.

## 4 Conclusion

In industrial product defect detection, defect data is not easy to obtain and a large number of defect data labeling wastes a lot of manpower and material resources. Therefore, the surface defect detection of industrial products based on a small number of labeled data is particularly important. In this paper, some methods for surface defect detection of industrial products based on a small number of labeled data are mainly introduced, and the methods are divided into two categories. The first category is traditional image processing-based industrial product surface defect detection. Training the model does not require a large number of labeled data, and it requires less time complexity when training the model, but this method has high requirements for the imaging environment of the image, and the method has poor adaptability. The second type is deep learning-based industrial product surface defect detection. This paper mainly introduces a deep learning-based industrial product surface defect detection methods suitable for a small number of labeled data. This method has strong adaptability, but the time complexity of this method is high. A large number of high-performance computing units are required. Therefore, how to improve the adaptability of the traditional image processing-based industrial surface defect detection method and how to reduce the time complexity of the deep learning-based industrial surface defect detection method suitable for a small number of labeled data and reduce the use of high-performance computing units is a problem that needs to be solved.